\documentclass{article}
 \usepackage[preprint, nonatbib]{neurips_2019}

\usepackage[utf8]{inputenc} % allow utf-8 input
\usepackage[T1]{fontenc}    % use 8-bit T1 fonts
\usepackage{hyperref}       % hyperlinks
\usepackage{url}            % simple URL typesetting
\usepackage{booktabs}       % professional-quality tables
\usepackage{amsfonts}       % blackboard math symbols
\usepackage{nicefrac}       % compact symbols for 1/2, etc.
\usepackage{microtype}      % microtypography
\usepackage{graphicx}
\usepackage{amsmath}
\usepackage{graphicx}
\usepackage{subcaption}
\usepackage{wrapfig}
\usepackage{xcolor}
\usepackage[ruled,vlined,noresetcount]{algorithm2e}
\newcommand{\eat}[1]{}
\def\hao{\textcolor{black}}

\def\zhu{\textcolor{black}}
\def\hui{\textcolor{black}}
\def\zfd{\textcolor{black}}

\newcommand\given[1][]{\:#1\vert\:}

\title{Continual Reinforcement Learning with Diversity Exploration and Adversarial Self-Correction}

\author{
	Fengda Zhu$^{\dagger}$, Xiaojun Chang$^{\dagger}$, Runhao Zeng$^{\ddagger}$, Mingkui Tan$^{\ddagger}$ \\
	$^{\dagger}$Monash University, $^{\ddagger}$South China University of Technology\\
	\texttt{zhufengdaaa@gmail.com; cxj273@gmail.com; } \\ \texttt{runhaozeng.cs@gmail.com; mingkuitan@scut.edu.cn}
}

\begin{document}
	
	\maketitle
	
	\begin{abstract}
		
		Deep reinforcement learning has made significant progress in the field of continuous control, such as physical control and autonomous driving. However, it is challenging for a reinforcement model to learn a policy for each task sequentially due to \emph{catastrophic forgetting}. Specifically, the model would forget knowledge it learned in the past when trained on a new task.
		We consider this challenge from two perspectives: 
		$i$) acquiring task-specific skills is difficult since task information and rewards are not highly related; 
		$ii$) learning knowledge from previous experience is difficult in continuous control domains. 
		In this paper, we introduce an end-to-end framework namely Continual Diversity Adversarial Network (CDAN). 
		We first develop an unsupervised diversity exploration method to learn task-specific skills using an unsupervised objective. Then, we propose an adversarial self-correction mechanism to learn knowledge by exploiting past experience. 
		The two learning procedures are presumably reciprocal. 
		To evaluate the proposed method, we propose a new continuous reinforcement learning environment named Continual Ant Maze (CAM) and a new metric termed Normalized Shorten Distance (NSD). The experimental results confirm the effectiveness of diversity exploration and self-correction. It is worthwhile noting that our final result outperforms baseline by 18.35\% in terms of NSD, and 0.61 according to the average reward.
		
	\end{abstract}
	
	\section{Introduction}
	
	Reinforcement learning has become increasingly popular due to its success in addressing challenging sequential decision-making problems~\cite{franccois2018introduction}. 
	Significant progress has been made in many real-world applications, such as playing video games~\cite{mnih2013playing,mnih2016asynchronous,booth2019marathon}, robot control~\cite{duan2016benchmarking, lillicrap2015continuous}, and robot navigation~\cite{jaderberg2016reinforcement, brodeur2017home, wu2018building,tai2017virtual}.
	\hao{Equipped with reinforcement learning, agents are able to learn strategies in very complex environments ~\cite{silver2017mastering,sun2018tstarbots} and even cooperate in multi-agent competition~\cite{bansal2017emergent}.}
	\zfd{However, unlike humans who can continually acquire skills and transfer previously learned knowledge throughout their life span, agents suffer from the \emph{catastrophic forgetting} issue. That is to say, reinforcement agents are prone to forget the knowledge they learned in the past.}
	
	To address the \emph{catastrophic forgetting} problem, continual reinforcement learning has been extensively studied. In this problem, 
	a reinforcement model continually learns over time by accommodating new knowledge while retaining previously learned skills (also called behaviors). Recently, Kaplanis \emph{et al.} \cite{kaplanis2018continual} propose a synaptic model equipped with tabular and deep Q-learning agent to solve this problem. However, when the number of tasks increases, the performance of this method is limited. 
	\zfd{More importantly, Q-learning becomes infeasible when applied to complex continuous control problems, where action spaces are continuous and often high-dimensional~\cite{sutton2018reinforcement, lillicrap2015continuous}}. 
	\zfd{In this paper, we focus on solving continual reinforcement learning problems in the field of continuous control, a task widely occurred in physical control~\cite{tassa2018deepmind} and autonomous driving~\cite{wang2018deep}.
	}
	One critical challenge of this task is that \zfd{rewards are not highly correlated to task information (\emph{e.g.}, task index and other configurations).} For example, in the early stages of training, a model would receive similar rewards, even \zfd{though} in different environments. 
	To solve this problem, researchers have proposed to learn different skills with respect to different contexts. 
	Eysenbach \emph{et al.} \cite{eysenbach2018diversity} \zfd{propose} an unsupervised exploration objective, \textit{diversity is all you need} (DIAYN), to learn diverse behaviors in a reinforcement learning paradigm. They \zfd{use} the classification probability from the discriminator as a reward, which\zfd{, however, } prevents the model from training in an end-to-end manner.
	\zfd{More critically, directly applying DIAYN to continual learning problems is infeasible. When} the discriminator becomes more confident, and rewards become \zfd{higher} in the latter stages, the model may abandon a good skill \zfd{simply} because it wants to behave differently. 
	%More critically, when the discriminator becomes more confident, and rewards become high in the latter stages, the model may abandon a good skill because it wants to behave differently. 
	\zfd{Recent works ~\cite{lopez2017gradient, riemer2018learning} introduce experience replay to learn from previous knowledge, which is able to tackle this limitation. } However, these methods only focus on independently and identically distributed sampled data, \zfd{making them impractical} in our non-stationary environment. \zfd{How to exploit the previous experience in the continuous control domain remains challenging. }

\begin{figure}
		\centering
		\includegraphics[width=0.8\linewidth]{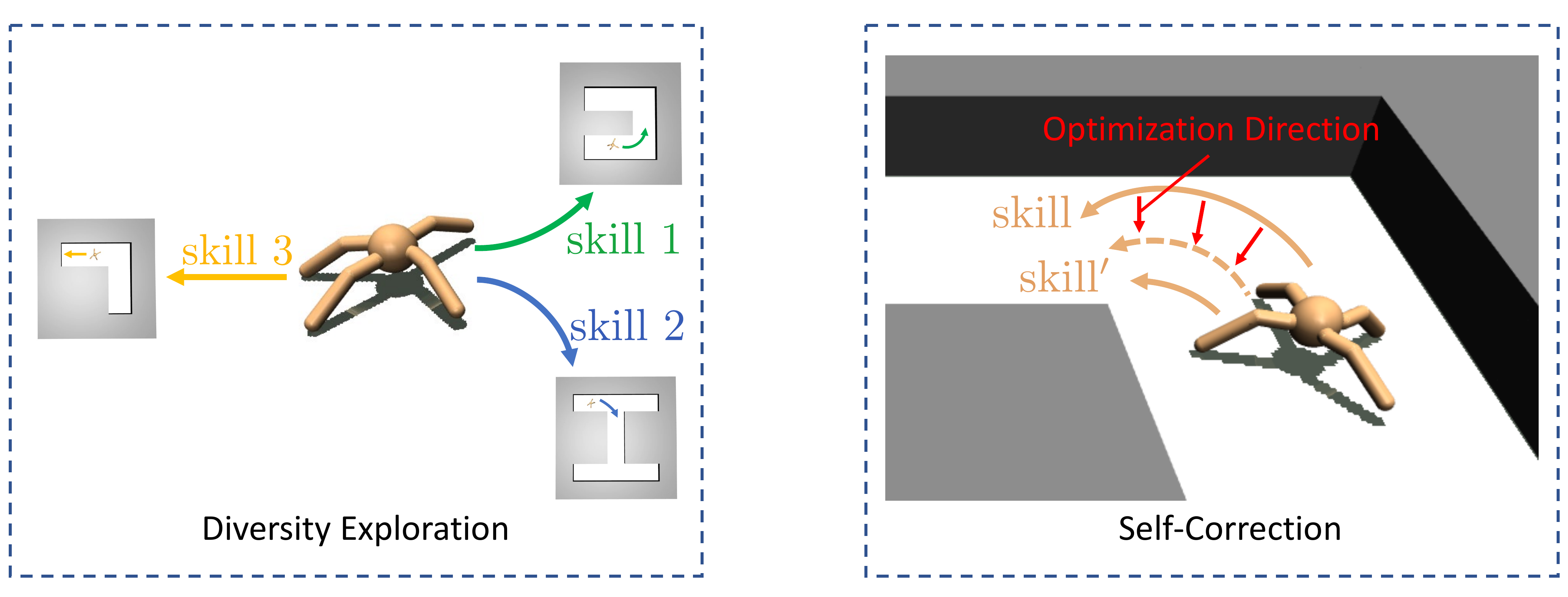}
		\caption{
			A simple demonstration of the proposed diversity exploration and self-correction. \textbf{Left}: With diversity exploration, our model learns to perform different skills with respect to different tasks. 
			\textbf{Right}: The "$\mathrm{skill}$" stands for the current skill, and the "$\mathrm{skill}{}'$" stands for previously learned skill. Model imitates $\mathrm{skill}{}'$ if previous skill performs better. 
		}
		\label{fig:demo}
		\vspace{-0.5cm}
	\end{figure}

	In this paper, we propose an end-to-end continual reinforcement learning framework in the domain of continuous control.
	As shown in Figure~\ref{fig:demo}, to address the above challenges, our method contains two parts: diversity exploration and self-correction.
	% Diversity exploration is an unsupervised learning process. Policy network and discriminator are jointly connected. Policy network minimizing the classification loss of discriminator to behave differently in order to find more useful skills. 
	\zfd{Diversity exploration is an unsupervised objective that optimizes a policy to learn task-specific skills.} \zfd{With the extra training signals provided by a discriminator, diversity exploration} encourages the policy to perform different behaviors with respect to task-specific contexts. To exploit the knowledge in previous experience, we explore an adversarial training mechanism to train the policy. In particular, we update the current policy using the previous experience when it forgets the \zfd{skills learned in the past}.
	In this way, the policy is optimized to maintain its performance in previous \hui{tasks}. To further enable the end-to-end training, our policy is designed to predict the next state \zfd{for the discriminator} in addition to the action. 
	
	% \hao{Diversity exploration optimizes policy to behave differently in order to find task-specific skills. Consequently, model learns to perform different behaviors with respect to task-specific contexts.
	%Self-correction is an adversarial mechanism to improve behavior by experience replay. We \textcolor{blue}{argue} that policy should be updated if the previous trajectory performs better than the trajectory sampled from the current policy. In these circumstances, the policy network will be optimized so as to behave more likely to former trajectories, which therefore makes discriminator harder to distinguish. We optimize policy by maximizing discriminated loss to achieve this. 
	% Self-correction updates policy if it performs worse than previous trajectory, which is well known as "catastrophic forgetting". In these circumstances, the policy network will be optimized to behave more similar to previous trajectories. 
	% Optimizing this objective will make discriminator harder to recover contexts from trajectories, which is an adversarial training procedure. 
	% To enable end-to-end training, Our policy predicts next state in addition to  action,  and discriminator recover context based on this predicted trajectory rather than real trajectory. 

	To demonstrate the performance of the proposed framework,
	we propose a new continual reinforcement learning setting called Continual Ant Maze (CAM). This environment instructs an ant to learn to navigate in multiple mazes. 
	% Even though the reward is informative and can be used as auxiliary indicators, we suggest an intuitive metric serves as a primary quantifier in comparing the performance of different models.
	Although the reward is widely used as an auxiliary indicator, we propose a more intuitive metric, \zfd{namely} normalized shorten distance (NSD), to better evaluate \zfd{the performance of each model} in our environment.
	% which is more suitable for performance evaluation 
	% We \textcolor{blue}{propose} a new metric called normalized shorten distance(NSD) to better evaluate our models. 
	Our experiments quantitatively and qualitatively evaluate how diversity exploration and self-correction help our model find more useful skills and avoid catastrophic forgetting. Experimental results suggest that our model outperforms baseline by 18.35\% NSD. 
	% footprint Such an significant improvement can also be seen in our demo. 
	
	% \begin{figure}
	% 	\centering
	% 	%\includegraphics[width=0.6\linewidth]{demo2.pdf}
	% 	\begin{subfigure}[b]{0.3\textwidth}
	% 		\includegraphics[width=\textwidth]{demo21.pdf}
	% 		\caption{Diversity Exploration}
	% 		\label{fig:exploration}
	% 	\end{subfigure}
	% \begin{subfigure}[b]{0.3\textwidth}
	% 	\includegraphics[width=\textwidth]{demo22.pdf}
	% 	\caption{Self-Correction}
	% 	\label{fig:self-correction}
	% \end{subfigure}
	% 	\caption{
	% 	A simple demonstration of exploration and self-correction. Solid line stands for current policy $\pi_\theta$. Dotted line in Figure~\ref{fig:exploration} represents a better policy find by exploration. And dotted line in Figure~\ref{fig:self-correction} demonstrates the self-correction process where policy $\pi_\theta$ improved by the supervision of trajectory sampled from policy before ($k<i$). Arrow points the direction of optimization. 
	% 	}
	% 	\label{fig:demo}
	% \end{figure}
	
	\section{Related Work}

	Our work is closely related to the following topics: deep reinforcement learning, unsupervised reinforcement learning, and continual learning.
	
	%\subsection{Deep Reinforcement Learning}
	\noindent\textbf{Deep Reinforcement Learning}
	%Reinforcement learning is a powerful tool for learning an agent making sequential decisions to maximize accumulative rewards.
	Research on deep reinforcement learning has been actively carried out due to
	its excellent performance in an Atari game via Deep Q-Network  (DQN) \cite{mnih2015human}. 
	Recently, many techniques \cite{mnih2016asynchronous,schaul2015prioritized,van2016deep,wang2015dueling} have been studied to improve the performance of the reinforcement learning algorithm. 
	\zfd{Even though Q-learning has achieved significant success, it is infeasible to apply Q-learning to continuous control problems due to the curse of dimensionality~\cite{duan2016benchmarking}. }
	%Despite the success achieved by Q-learning, it is unfeasible to use it in a continuous control problem due to the curse of dimensionality~\cite{duan2016benchmarking}. 
	In contrast, on-policy algorithms like A3C~\cite{mnih2016asynchronous} can overcome this limitation because of their capability of dealing with continuous and high-dimensional inputs. However, A3C suffers from poor data efficiency and robustness. To solve this issue, TRPO \cite{schulman2015trust} optimizes policy under certain constraints to guarantee monotonic improvement. \zfd{Nevertheless}, TRPO is too complicated and is incompatible with architectures that have auxiliary losses and shared weights. This drawback is further tackled by PPO \cite{schulman2017proximal} using a clipped surrogate objective. 
	%Nowadays, reinforcement learning algorithm is often applied in Atari games or robotics problems, and it also has many potential applications in computer vision fields. 
	
	%Techniques such as prioritized experience replay \cite{schaul2015prioritized}, double DQN \cite{van2016deep}, dueling DQN \cite{wang2015dueling}, and A3C \cite{mnih2016asynchronous} have been studied to improve the performance of the reinforcement learning algorithm. 
	
	%In continuous control problem, Q-learning quickly is infeasible due to the curse of dimensionality [xxx]. In contrast, on policy algorithms like A3C~\cite{mnih2016asynchronous} and PPO 
	
	%succeed due to they work on continuous and high-dimensional inputs. 
	
	%Reinforcement learning is originated from the psychological and neuroscientific understandings of how humans learn to optimize
	%their behaviors in an environment. It can be mathematically
	%formulated as a Markov decision process  (MDP). With
	%a person being generalized to an agent, the behaviors being generalized to a set of actions, a typical reinforcement
	%learning problem can be formulated as an agent optimizes
	%its policy of actions by maximizing the numerical rewards
	%it receives from an environment.
	%
	%Q learning: discrete action space
	%
	%Policy gradient: PPO

	%\subsection{Unsupervised Reinforcement Learning}
	\noindent\textbf{Unsupervised Reinforcement Learning} Recently, several researchers have proposed unsupervised training objectives for learning diverse skills. 
	%based on their distinctiveness. 
	Gregor \emph{et al.} \cite{gregor2016variational} introduce a formalism of intrinsic control maximization for unsupervised option
	learning. They first define options as policies with a termination condition and propose to discover diverse intrinsic options using an information theoretic learning criterion. 
	Eysenbach \emph{et al.} \cite{eysenbach2018diversity} \zfd{optimize} an information theoretic objective with a maximum entropy policy. In this way, the agent learns skills that explore large parts of the state space and ensures that each skill is individually distinct.
	%Achiam \emph{et al.} \cite{achiam2018variational} discuss a connection between variational option discovery and variational auto-encoders. By solving a max-max optimization problem, the context vector is associated with trajectories, and each context vector thus corresponds to a distinct option. 
	Achiam \emph{et al.} \cite{achiam2018variational} discuss a connection between variational option discovery and variational auto-encoders by solving a max-max optimization problem. 
	Our approach is closely related to \cite{eysenbach2018diversity, achiam2018variational} but distinct from it. 
	In addition to the max-max unsupervised optimization function, we propose a min-max adversarial  manner to \zfd{learn from previous experience}. 
	%Instead of solving the max-max optimization problem, we propose to play the min-max game in an adversarial manner.

	%\textbf{Diversity is All You Need: Learning Skills without a Reward Function}
	%
	%They: unsupervised learning
	%
	%Us: add unsupervised information supervised training period
	
	%\textbf{Variational Option Discovery Algorithms}
	%
	%They: Max diversity
	%
	%Us: memory to prevent forget, max diversity, and imitate good skill
	%
	%They: max max problem
	%
	%Us: in addition to max max problem, we reuse decoder to perform an adversarial min max problem

	%\subsection{Continual Learning}
	\noindent\textbf{Continual Learning} Continual learning is a long-standing goal of machine learning, where agents learn a series of tasks experienced in sequence.
	%Reinforcement learning  (RL) agents, while able to achieve human-level performance in
	%complex games like Go, usually focus on becoming really proficient at one task, and train from scratch in each new problem they face. In contrast, humans acquire skills and build on them to solve increasingly complex tasks. 
	%Recently, continual learning has attracted much attention \cite{bengio2009curriculum,fernando2017pathnet,lopez2017gradient,riemer2018learning,rusu2016progressive}. 
	Inspired by the human education system, Bengio \emph{et al.} \cite{bengio2009curriculum} propose a continuation method, namely curriculum learning, to solve complex sequences of tasks. A key issue of this paradigm is the catastrophic forgetting, \emph{i.e.}, the model often forgets the knowledge learned from previous tasks. To tackle this problem, Rusu \emph{et al.} \cite{rusu2016progressive} propose progressive networks. They instantiate a new neural network for solving each task while enabling transferring features of previously learned networks. Another attempt along this direction is the PathNet. 
	Fernando \emph{et al.} \cite{fernando2017pathnet} use a generic algorithm to select a pathway through a neural network. The pathway selected in the previous task is fixed, and the agent is trained to select the best pathway for other tasks. 
	%To monitor the performance of continual learning methods and assess the ability of the learner, Lopez-Paz \emph{et al.} \cite{lopez2017gradient} design three evaluation metrics. 
	Lopez-Paz \emph{et al.} \cite{lopez2017gradient} propose to use an episode memory and inequality constraints to enable the positive backward transfer. Inspired by \cite{lopez2017gradient}, Riemer \emph{et al.} \cite{riemer2018learning} encourage the network to share parameters when gradient directions align and keep parameters separate when gradients cause interference in opposite directions.

	%Fernando \emph{et al.} \cite{fernando2017pathnet} use a generic algorithm to select the best pathway through a neural network. Once the previous task has been trained, the best fit pathway is fixed, which means its parameters are no longer allowed to change. All other parameters not in an optimal path are reinitialized and the agent is trained to select the best pathway for the consequent task. 

	%simpler training data are fed into the network first and then the more difficult dataset. This method can guide the training process to converge faster and reach better minima.

	%\textbf{Distral: Robust Multitask Reinforcement Learning}
	%
	%They: multitask step wise distillation
	%
	%Us: consider episode wise constrain. Different from distillation, we use adversarial training to imitate behaviors. 
	%
	%
	%\textbf{Learning to Learn without Forgetting By Maximizing Transfer and Minimizing Interference}
	%
	%They: off policy, no exploration, step wise constrain
	%
	%Us: on policy, exploration, episode with constrain
	
	\section{Proposed Method}
	
	% \hao{Problem-->Challenge-->Solution}
	In our work, we study continual reinforcement learning in a continuous control environment. The environment consists of several scenes, e.g., mazes in AntMaze, house scenes in Indoor Room Navigation, etc. 
	The configurations of these scenes are different in size, shape, and complexity, thus leading to different reward functions. 
	In our setting, an agent is required to learn from multiple tasks, \emph{e.g.}, from task $T_1$ to task $T_N$, where $N$ stands for the number of tasks. There are two challenges to overcome this problem. First, the reward and policy gradient loss are not highly related to the task information, resulting in insufficient training signals for learning task-specific skills.
	% One critical challenge of this problem is that the reinforcement models are prone to forget the knowledge learned previously, known as ``catastrophic forgetting''. 
	Second, in continuous control tasks, 
	the previous experience is high-dimensional and noisy, making it challenging to leverage the previous knowledge.

	% Therefore, the state space, the probabilistic state-action transition function, and the reward function can also be different. 
	% The problem of continually learning a policy which is able to perform on arbitrary tasks is challenging. 
	%Though the actions space and probabilistic action-state transition function is shared among scenes, other configurations of these scenes are different. Information such as size, shape, and complexity of each scene are different. 
	
	To address the above challenges, we propose a framework called Continual Diversity Adversarial Network (CDAN).  Figure~\ref{fig:demo} shows a high-level explanation of how our method works.
	We first propose diversity exploration to solve the challenge that reward function is not highly correlated to the task information. It is an unsupervised method that encourages the model to explore possible skills and to learn relationships between task scenes and policy skills. 
	\zfd{To prevent the model from forgetting skills, we propose an adversarial learning method called self-correction to exploit previous experience. }
	In our work, the experience is a trajectory sampled from the reinforcement model. If the previous trajectory outperforms the trajectory of the current model, the model imitates the previous trajectory by maximizing the discriminated loss. 
	
	% Our framework consists of two parts, diversity exploration, and self-correction. 
	% Diversity exploration is an unsupervised method that encourages the model to explore possible skills and learns a relationship between task scenes and policy skills. 
	% As a result, the model learns to perform a particular mode of behavior for each scene. 
	% To avoid model "forget" skills learned in the latter stages, we propose an adversarial transfer learning method called self-correction to exploit previous experience. If previous experience outperforms skills of the current model, the model imitates the previous trajectory by maximizing the discriminated loss. 
	
	\begin{figure}[t]
		\centering
		\includegraphics[width=0.95\linewidth]{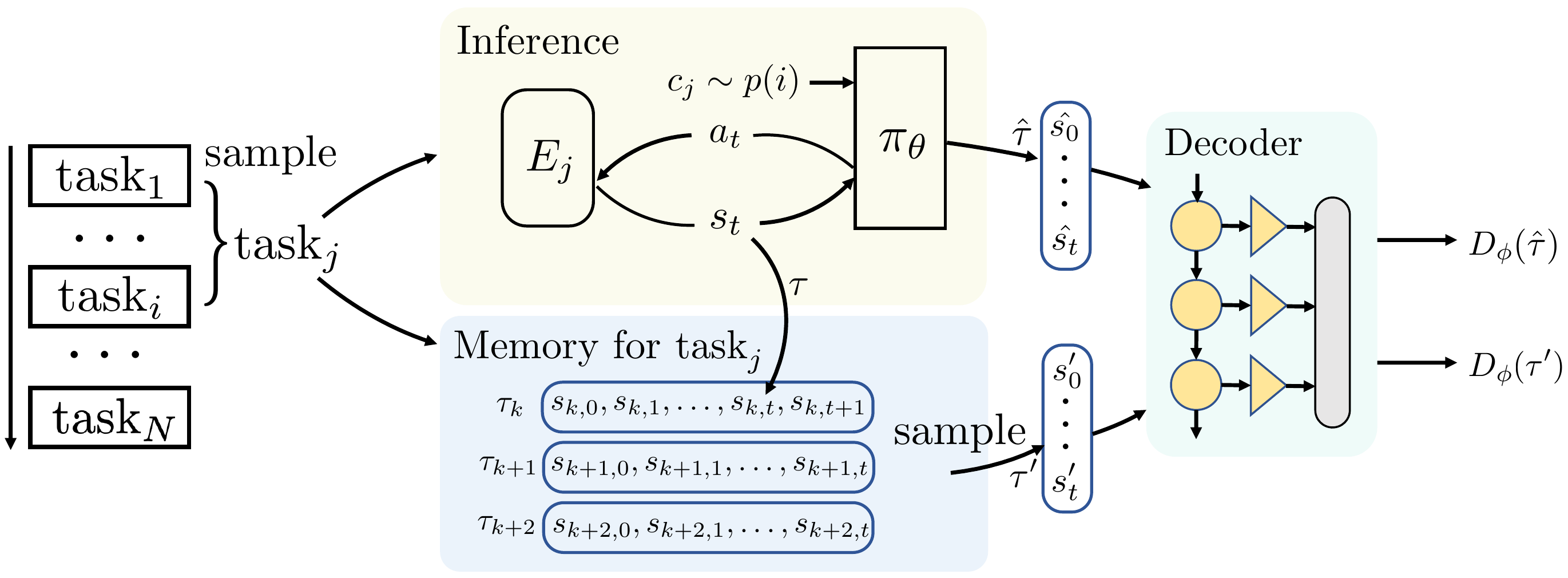}
		\caption{
			The pipeline of the Continual Diversity Adversarial Network  (CDAN). The current step of training is $i$, and  context distribution function over tasks is $p (i)$. Here, $c_j$ is a vector of context feature sampled from $p (i)$, which belongs to $\mathrm{task}_j$. 
			$D_\phi$ consists of an LSTM layer(yellow circles), a fully connected layer(yellow triangles), and an average pooling layer (grey rectangle). 
		}
		\label{fig:pipeline}
		\vspace{-0.4cm}
	\end{figure}
	
	\subsection{Preliminaries}
	
	For convenience, we formulate each task $T_i$ as a Partially Observable Markov Decision Process (POMDP), defined by $T_i= (S, A, O, P, R)$. Here, $s_t \in S$ describes the state of an agent at time $t$, and $a_t \in A$ is the action performed by the agent under the condition of observation $o_t \in O$. The probabilistic state-action transition function is represented as $p (s_{t+1} \given s_{t}, a_{t})$. The reward $r_t = R(s_{t}, a_{t})$ is a predefined function related to the task. \zfd{Note that for simplify notation without loss of generality}, we use $s_t$ to represent what the agent observes from this task. 
	
	The policy function $\pi$, defined by $\theta$,  \zfd{receives a state $s_{t}$ at step $t$ as input} and predicts an action distribution. The action $a_{t}$ is sampled from the action distribution, marked as $a_{t} \sim \pi_{\theta} (s_{t})$. 
	%For each task, start state pairs $ (t_i, s_0)$, $\pi$ generates an episode $ep_k$ consists of start, action, reward pairs  ($s_{t}$, $a_{t}, r_{t}$). 
	Starting from state $s_0$, \zfd{$\pi_\theta$} generates a state trajectory $\tau= (s_0, \dots  s_t)$. We optimize the probability of action with the greatest advantage via policy gradient objective:
	\begin{equation}
	\underset{\theta}{\mathrm{max}}\  \mathbb{E}_{a_t, s_t \sim \pi (\theta)}\  \mathrm{log}\left (\pi_{\theta} (a_{t}\given s_{t})\right)A (a_{t},s_{t}),
	\label{equation:rl}
	\end{equation}
	where $A (a_{t},s_{t}) = Q (a_{t},s_{t})-V (s_{t})$ means the advantage of choosing an action $a_{t}$. $Q (a_{t},s_{t})$ stands for the value choosing $a_{t}$ with state $s_{t}$ and $V (s_{t})$ stands for the accumulated value of state $s_{t}$. 
	
	% But A3C suffers poor data efficiency and robustness. Thus, PPO propose a clipped surrogate objective to address these challenges
	
	%\begin{equation}
	%r_{t} (\theta) = \frac{\pi_{\theta} (a_t \given s_t)}{\pi_{\theta_{old}} (a_t \given s_t)}
	%\end{equation}
	
	%\begin{equation}
	%J^{CLIP} (\theta) = arg\  \underset{\theta}{max}\  \mathbb{E}_{a_t, s_t \sim \pi (\theta)}\  min (r_t (\theta)A (a_t, s_t), clip (r_t (\theta), 1- \varepsilon , 1+ \varepsilon )A (a_t, s_t))
	%\end{equation}
	
	%Here $r_t$ is the ratio of the probability under the new and old policies and $ \varepsilon $ is a hyperparameter of clipping. Compared with A3C, model trained on PPO algorithm is more robust and further enable our continual experiments. 
	
	\subsection{End-to-end Diversity Exploration} \label{sec:de}
	
	% Our experiments shows that even though task information is accorporated into the input for agent, model still can collapse. 
	\zhu{
		Since task information does not have a high correlation with reward and policy gradient loss, models may collapse even though task information is incorporated into the input.
		To be specific, model lacks of useful training signals \zfd{since} it receives similar rewards in different tasks. 
		As a result, learning task-specific skills is extremely difficult. 
		To address this issue, we seek to provide an extra training signals for model to learn the relationship between skills and tasks. 
	}
	% Thus, models are likely to confuse policy strategies for different tasks. 

	We propose a diversity exploration method to learn conditioned skills. 
	% so that the model behaves differently with respect to different tasks. 
	%We formulate our learning approach of exploration as follows. 
	We define context $c \sim p (T_i)$, where $c$ contains task-specific information and a random variable, and $p(T_i)$ is a context distribution. 
	\zfd{Note that even with different random variables,  different $c$ belong to the same task as long as their task information is the same.}
	We concatenate $c$ together with $s_t$  and feed it to policy $\pi$ for each time step $t$. 
	%Different from DIAYN and VALOR, which optimize model by real trajectory $\tau$, 
	%we use a approximate $\hat{\tau}= (\hat{s_0}, \dots, \hat{s_n})$ estimated by policy $\pi (a_t , \hat{s_t} \given c)$.
	% Different from DIAYN
	%and VALOR introduce VALOR
	Our policy $\pi_\theta$ predicts next step $\hat{s_{t}}$ in addition to action $a_t$, represented as $\pi_\theta(\hat{s_t}, a_t \given s_{t-1}, c)$. 
	By interacting with the environment, our model samples a real trajectory $\tau= (s_0, \dots, s_n)$ and a predicted trajectory $\hat{\tau}=(\hat{s_0}, \dots, \hat{s_n})$. 
	Rather than identifying behavior \zfd{with} an arbitrary state $s_t$, we use a discriminator $D_\phi$ to recover a context from a complete trajectory. 
	% \hao{We use the following objective to train the discriminator:}
	We use the following objective to optimize policy $\pi_\theta$ and discriminator $D_\phi$ jointly:
	\begin{equation}
	\underset{\theta, \phi}{\mathrm{max}}\   \mathbb{E}_{c, \hat{\tau} \sim \pi}\  \mathrm{log} D_\phi (\pi_\theta (\hat{\tau} \given c)).
	\label{equation:joint}
	\end{equation}
	Here, the discriminator $D_\phi$ learns to classify $c$ for different trajectories $\tau$. 
	Also, the policy $\pi_\theta$ learns to generate similar trajectories for the same context $c$ and different trajectories for different contexts in order to facilitate the classification of discriminator on the other hand. The major advantage of this formulation is we can optimize policy \zfd{$\pi_\theta$} and discriminator \zfd{$D_\phi$} in an end-to-end manner.

	\zfd{Note that we use states rather than actions to distinguish skills since actions are not observable.}
	We regularize each state $\hat{s_t}$ in $\hat{\tau}$ with L1 loss:
	\begin{equation}
	\underset{\theta}{\mathrm{min}}\  \mathbb{E}_{\tau, \hat{\tau} \sim \pi, c}\ \sum_{t=0}^{n}|\pi_\theta (\hat{s_t} \given s_{t-1}, c) - s_t|.
	\label{equation:l1}
	\end{equation}
	% Here, $s_t$ is the state in real trajectory $\tau$ and $\hat{s_t}$ is the state at time step $t$ given $s_{t-1}$ and $c$. 
	This objective serves as a constraint guaranteeing that the discriminator uses states rather than actions to distinguish skills.
	Also, it is an auxiliary function introducing additional supervision to improve data efficiency and help model converge faster. 
	
	\subsection{Adversarial Self-Correction}
	% \subsection{Adversarial Self-Correction for Learning Previous Knowledge}

	\zhu{In continual reinforcement learning, models are prone to forget previously learned skills when trained on new tasks. Experience replay is a widely adopted solution to this problem.
		% Experience replay exploits previous experience to solve this problem. 
		In continuous control, however, since trajectories are high-dimensional and noisy, exploiting trajectories to learn previous knowledge is extremely hard. 
		% The discriminator described in \zfd{section}~\ref{sec:de} is able to recover contexts from trajectories, which contain rich information about skills. 
		\zfd{The discriminator described in \zfd{section}~\ref{sec:de} is able to recover contexts from trajectories. 
			Therefore, the context prediction $D(\hat{\tau})$ contains rich information about skills, 
			for which we consider using $D(\hat{\tau})$ to leverage knowledge learned previously. 
		}
	}
	
	\begin{algorithm}[!tb]
		\KwIn{$E$: an environment containing multiple \zfd{tasks}}
		\KwIn{\zfd{$p (T_i)$: a context distribution defined by tasks $T_i$}}
		\KwIn{$\alpha_1, \alpha_2, \alpha_3$: learning rates}
		%\KwResult{how to write algorithm with \LaTeX2e }
		$M \leftarrow \{\}$\\
		\For  {$i = 1,\dots, T$}{
			\While  {$j =1,\dots, \zfd{\mathrm{time\ steps}}$}{
				$\tau \leftarrow \{\}$, $\tau_{c_j} \leftarrow \{\}$\\
				Sample batch of \zfd{task contexts} $c_j \leftarrow \zfd{p (T_i)}$ \\
				\ForEach{$c_j$}{
					%$s_0 \leftarrow E.reset (c_j)$ \\
					%$s_0 \leftarrow \mathrm{initial}\ E (c_j)$ \\
					\zfd{$s_0 \leftarrow \mathrm{initial}(E,c_j)$ // Initialize environment with context}\\ 
					// Sample trajectories, trajectory predictions, discounted rewards, values, actions \\
					$\tau_{c_j}, \hat{\tau_{c_j}}, R, V, A = \mathrm{inference} (\pi, s_0, E)$ \\ 
					$\theta \leftarrow \mathrm{Adam} (R, V, A, \tau_{c_j}, \hat{\tau_{c_j}}, \theta, \phi, \alpha_1)$ // Optimize $\pi_\theta$ by \zfd{ Formulation~(\ref{equation:rl}),~(\ref{equation:joint}),~(\ref{equation:l1})} \\
					$\phi \leftarrow \mathrm{Adam} (c_j, \hat{\tau_{c_j}}, \phi, \alpha_2)$ // Optimize $D_\phi$ by Formulation~(\ref{equation:joint}) \\
					$\tau \leftarrow \tau \cup \{\tau_{c_j}\}, \hat{\tau} \leftarrow \hat{\tau} \cup \{\hat{\tau_{c_j}}\}$ \\
				}
				% 			// Collect batch of trajectories \\
				% 			$\tau \leftarrow \{\tau_{c_j}, \dots \},  \hat{\tau} \leftarrow \{\hat{\tau_{c_j}}, \dots \}$ \\
				$M \leftarrow M \cup \{\tau\}$ \\ 
				$\tau{}' \leftarrow \mathrm{sample} (M)$  // Sample batch of trajectories from memory \\
				$\theta, \phi \leftarrow \mathrm{Adam} (\hat{\tau}, \tau{}' ,\theta, \phi, \alpha_3)$ // Jointly Optimize $\pi_\theta$ and $D_\phi$ by Formulation~(\ref{equation:adversarial}) \\
			}
		}
		\caption{Continual Deep Reinforcement Learning}
		\label{alg:all}
	\end{algorithm}
	
	%As shown in Figure~\ref{fig:pipeline}, we introduce our pipeline. 
	%we introduce a memory $M$ to store trajectories generated by the current policy $\pi_\theta$ for each step. 
	%To exploit $D(\hat{\tau})$, we propose our pipeline shown in Figure~\ref{fig:pipeline}. 
	The method we use to exploit $D(\hat{\tau})$ is shown in Figure~\ref{fig:pipeline}. 
	When training the model in $\mathrm{task}_j$ ($\mathrm{task}_j$ can be $\mathrm{task}_i$ or \zfd{a} task prior to $\mathrm{task}_i$), we sample trajectories \zfd{by inference} and store them in memory $M$. \zfd{Then}, we sample a batch of trajectories from $M$. 
	%\zfd{Note that each $\tau{}'$ sampled from $M$ corresponds to each $\hat{\tau}$ sampled from the current policy respectively.} 
	We define $\tau{}'$ as a trajectory sampled from $M$ and $\hat{\tau}$ as a trajectory sampled from the current policy $\pi_\theta$. 
	Note that we ensure that $\hat{\tau}$ and $\tau{}'$ are sampled from same \zfd{task, with the same starting position and the same pose}. 
	Self-correction compares a trajectory $\tau$ from the current policy $\pi_\theta$ with a trajectory $\tau{}'$ sampled from $M$. 
	%generated by a previous policy $\pi_{\theta{}'}$ under the same environment, 
	We want $\hat{\tau}$ to be closer to $\tau{}'$ if $\tau{}'$ performs better (\emph{e.g.}, the accumulated reward is greater) but far away from $\tau{}'$ elsewise (optimized by Formulation~(\ref{equation:joint})). 
	To minimize the distance of two skills, we utilize the trajectory information decoded by the discriminator. 
	To be specified, we use the context recovered by discriminator, $D_\phi (\tau{}')$ , to serve as a soft label of imitation. 
	On the other hand, we optimize $D_\phi$ to distinguish $\hat{\tau}$ and $\tau{}'$. 
	\zfd{Thus, we propose the following minimax formulation, where $\pi_\theta$ and $D_\phi$ are jointly optimized:} 
	\begin{equation}
	\underset{\phi}{\mathrm{min}}\ \underset{\theta}{\mathrm{max}}\  \mathbb{E}_{c, \hat{\tau} \sim \pi, \tau{}' \sim M}
	D_\phi (\tau{}')\mathrm{log} D_\phi (\pi_\theta (\hat{\tau} \given c)). 
	\mbox{\ \ \ \ \ if \   $\sum_{i=0}^{t}  \gamma^{i} r{}'_{i} > \sum_{i=0}^{t} \gamma^{i} r_i$}.
	%&\text{where $\lambda_{1}+\lambda_{2} = 1$} \notag
	\label{equation:adversarial}
	\end{equation}
	\zfd{We use discounted reward to indicate the performance of trajectories and denote $\gamma$ for the discount factor. }
	\zhu{
		Note that Formulation~(\ref{equation:adversarial}) is essentially the minimax problem, which optimizes $\pi_\theta$ and $D_\phi$ alternatively in GAN~\cite{NIPS2014_5423} manner. 
		The only difference is that Formulation~(\ref{equation:adversarial}) is optimized when condition $\sum_{i=0}^{t}  \gamma^{i} r{}'_{i} > \sum_{i=0}^{t} \gamma^{i} r_i$ is satisfied. 
		In this way, our model is optimized by its previous knowledge without additional supervision. 
		Thus we call this method Adversarial Self-Correction(ASC). 
	}
	Generally, $\pi_\theta$ is optimized to make $D_\phi$ more difficult to distinguish $\hat{\tau}$ and $\tau{}'$, while $D_\phi$ is optimized adversarially to recover contexts from the trajectories correctly. 
	%Note that the context of $\hat{\tau}$ is hardly to be equal to the context of $\tau{}'$ since contexts contain random variable. 
	We further provide details for CDAN in Algorithm~\ref{alg:all}. 
	
	\begin{figure}[bt]
		\centering
		\includegraphics[width=0.9\linewidth]{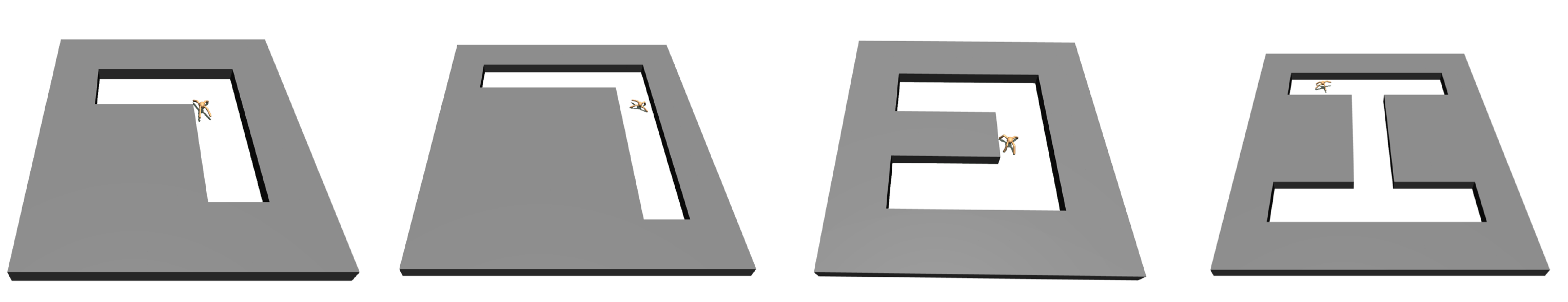}
		\caption{
			Mazes in our dataset. The mazes have different size, shape, and complexity.
			% 		We shows different size and complexity of mazes in our dataset. 
		}
		\label{fig:mazes}
		\vspace{-0.4cm}
	\end{figure}
	
	\section{Experiments} % \subsection{Continual Environment} 
	\subsection{Experiment Setup}
	
	Our environment is complex since it combines continuous control, maze navigation, and continual learning. 
	Figure~\ref{fig:mazes} shows 4 mazes in the Continual Ant Maze environment with different sizes. 
	\zfd{In each task, a robot starts from an fixed initial point with a random pose, being instructed to navigate to a fixed goal. }

	To train the reinforcement model, Duan \emph{et al.} \cite{duan2016benchmarking} propose a sparse reward function, which is 1 if the agent reaches the goal and 0 elsewise.
	However, this reward function is impractical because it cannot provide sufficient training signals. To address this problem, we propose a denser reward function as follows:
	% Different from previous work~\cite{duan2016benchmarking} provides a sparse reward function which is 1 if the agent reach goal and 0 elsewise, we propose a denser reward function as follows:
	\begin{equation}
	r =  d (p_t, g) - d (p_{t+1}, g) - \eta , 
	\end{equation}
	where $r$ is per step reward, $g$ is the position of the goal. $d (p_t, g)$ stands for the shortest distance from point $p_t$ at time $t$ to $g$, and $\eta$ is a time punishment term encourages model to reach the goal quicker. This reward function measures how much agent shorten its distance to the goal for every step. Experiments show that it is facilitate training since the agent receives a clear optimization direction even though it is far away from the goal at the early stage.

	% Task encoding $c$ is sampled following a uniform distribution, $c \sim $Uniform ($i$), where i is the task id of current task.
	Most of the works regard reinforcement learning problems as reward maximization problems, where the reward also serves as the primary indicator of performance. As demonstrated in ~\cite{duan2016benchmarking}, for example, the evaluation metric is defined as the mean reward of the whole trajectory. In this work, however, we propose a more intuitive metric considering every tasks to reflect the performance of each model. 
	%This measure ensures the model considers all of the previous tasks rather than overfit on the latest task $i$. 
	In testing, a robot starts at the same position with random poses. The robot stops if it touches the goal area or time limit exceeds. Shorten distance(SD) are calculated for each tasks by how much distance the robot shortens at the end of each episode. Overall, we propose a metric called normalized shorten distance (NSD) to evaluate the performances of each model, 
	\begin{equation}
	\mathrm{NSD} = \frac{1}{NT} \sum_{i=1}^{N} \sum_{j=1}^{T} \frac{d (p_{0}^{i}, g^{i}) - d (p_{n}^{i,j}, g^{i})}{d (p_{0}^{i}, g^{i})}. 
	\end{equation}
	$N$ stands for the number of tasks, and $T$ stands for the number of trajectories. Goal point $g^i$ and 
	start point $p_0^i$ do not change over trajectories for the same tasks. $p_{n}^{i,j}$ stands for the last position of trajectory $j$.

	\noindent\textbf{Implementation Details} Motivated by recent research~\cite{achiam2018variational}, we train a model from easy mazes to hard mazes incrementally. We use PPO algorithm to implement our baseline since it is more robust and more data efficient compared with A3C. Baseline model concatenates task information $c$ with every state $s_t$ and then feeds it to policy network as input. 
	
	In our implementation, we sample trajectories with a maximum length of 2048. $\pi_\theta$ is optimized by PPO with batch size 32 and learning rate 0.001. The discounted reward factor $\gamma$ is 0.99 and clip range is 0.2. We fix the length of discriminator input to 100 and the length of self-correction to  100 as well. All models are trained for $1e6$ time steps in total. Every training process costs 2 days of experience on a Titan X GPU device.

	\subsection{Results and Discussions}
	
	In this section, we focus on answering the following questions.  ($i$) Does our baseline be sufficiently trained.  ($ii$) Is diversity exploration effective and be helpful to solve catastrophic forgetting.  ($iii$) How do diversity exploration and self-correction benefit with each other and improve performance.

	\begin{figure}
		\centering
		\begin{subfigure}[b]{0.32\textwidth}
			\includegraphics[width=\textwidth]{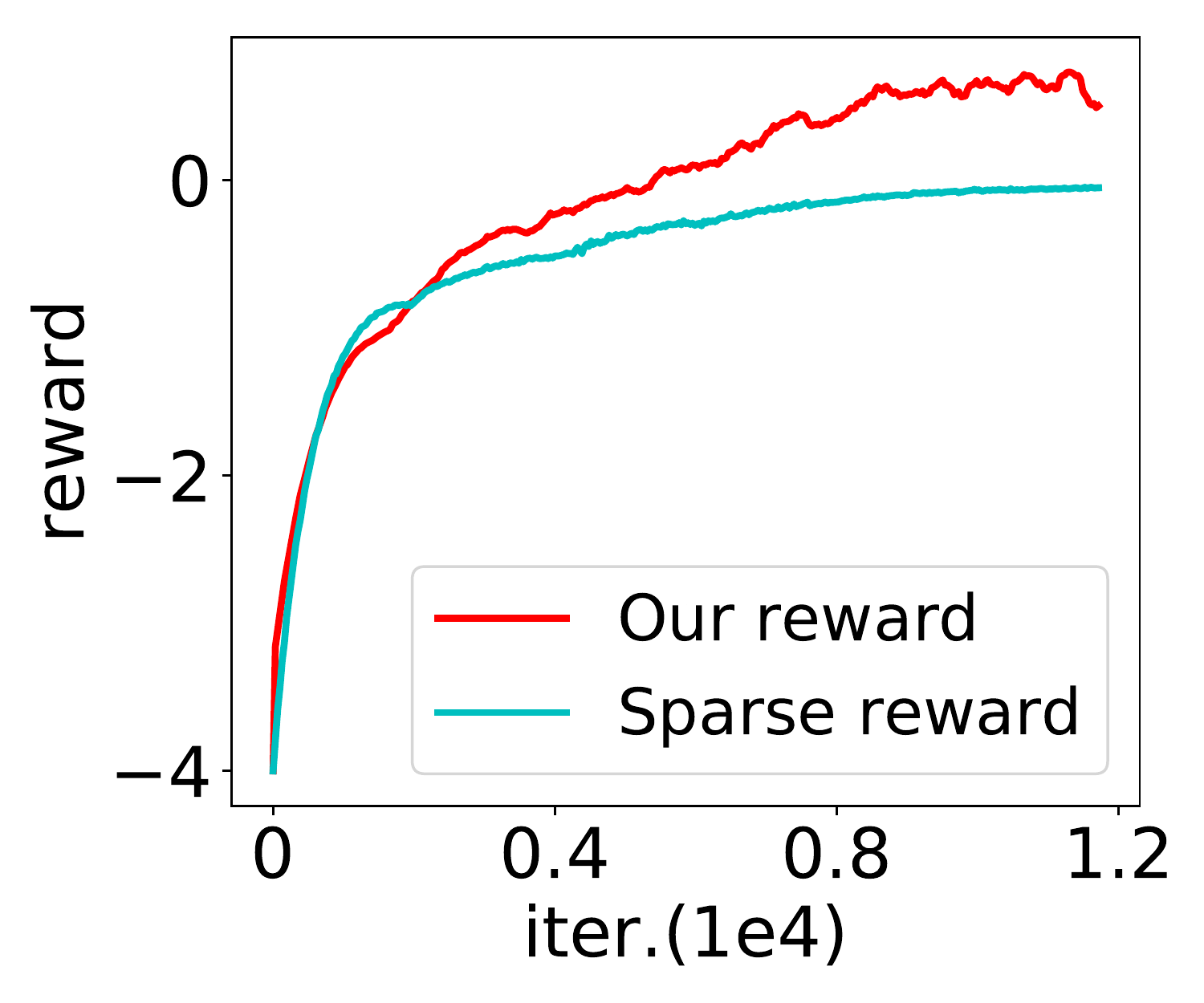}
			\caption{Reward Functions}
			\label{fig:rewards}
		\end{subfigure}
		\begin{subfigure}[b]{0.31\textwidth}
			\includegraphics[width=\textwidth]{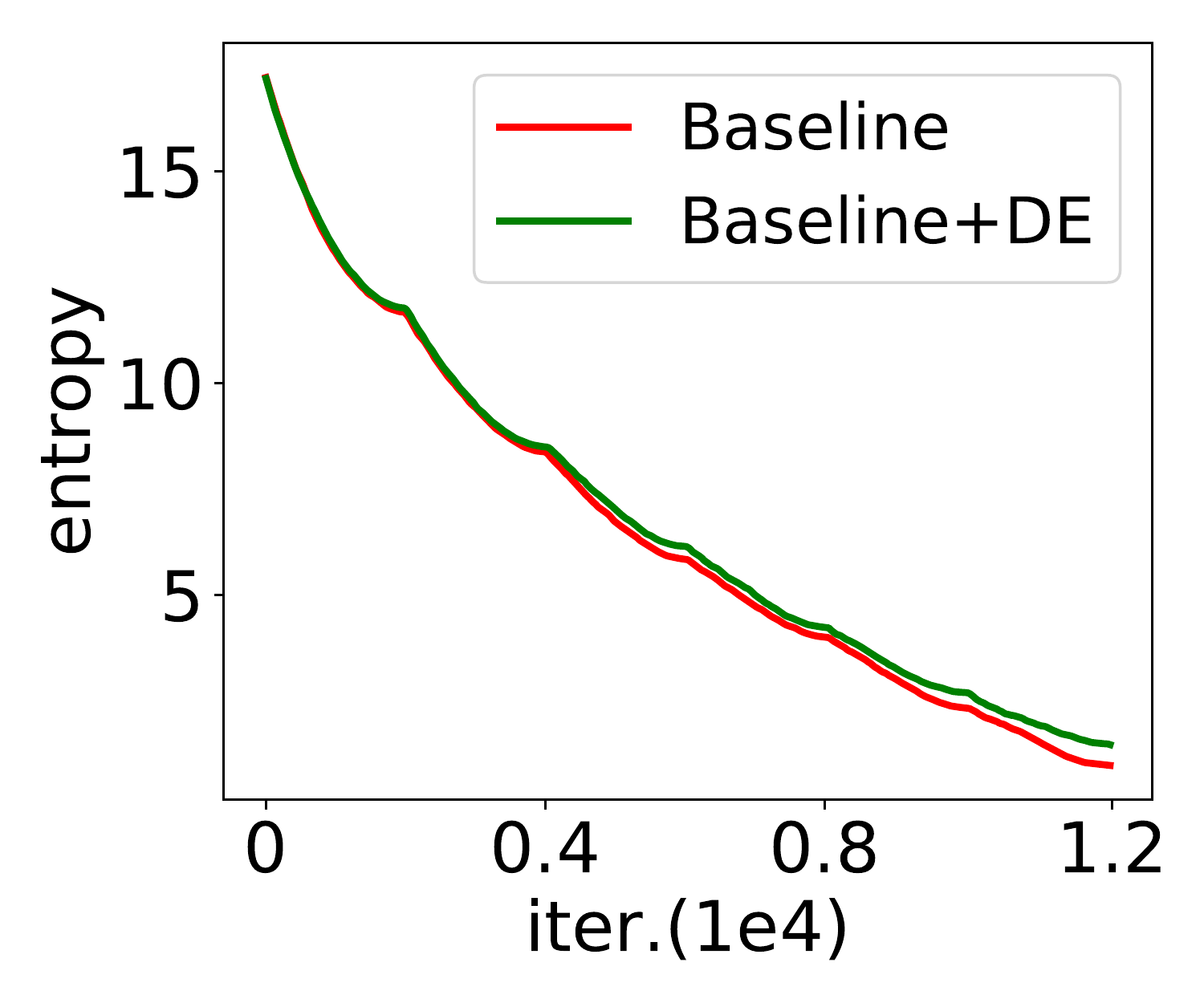}
			\caption{Policy Entropy}
			\label{fig:entropy}
		\end{subfigure}
		\begin{subfigure}[b]{0.31\textwidth}
			\includegraphics[width=\textwidth]{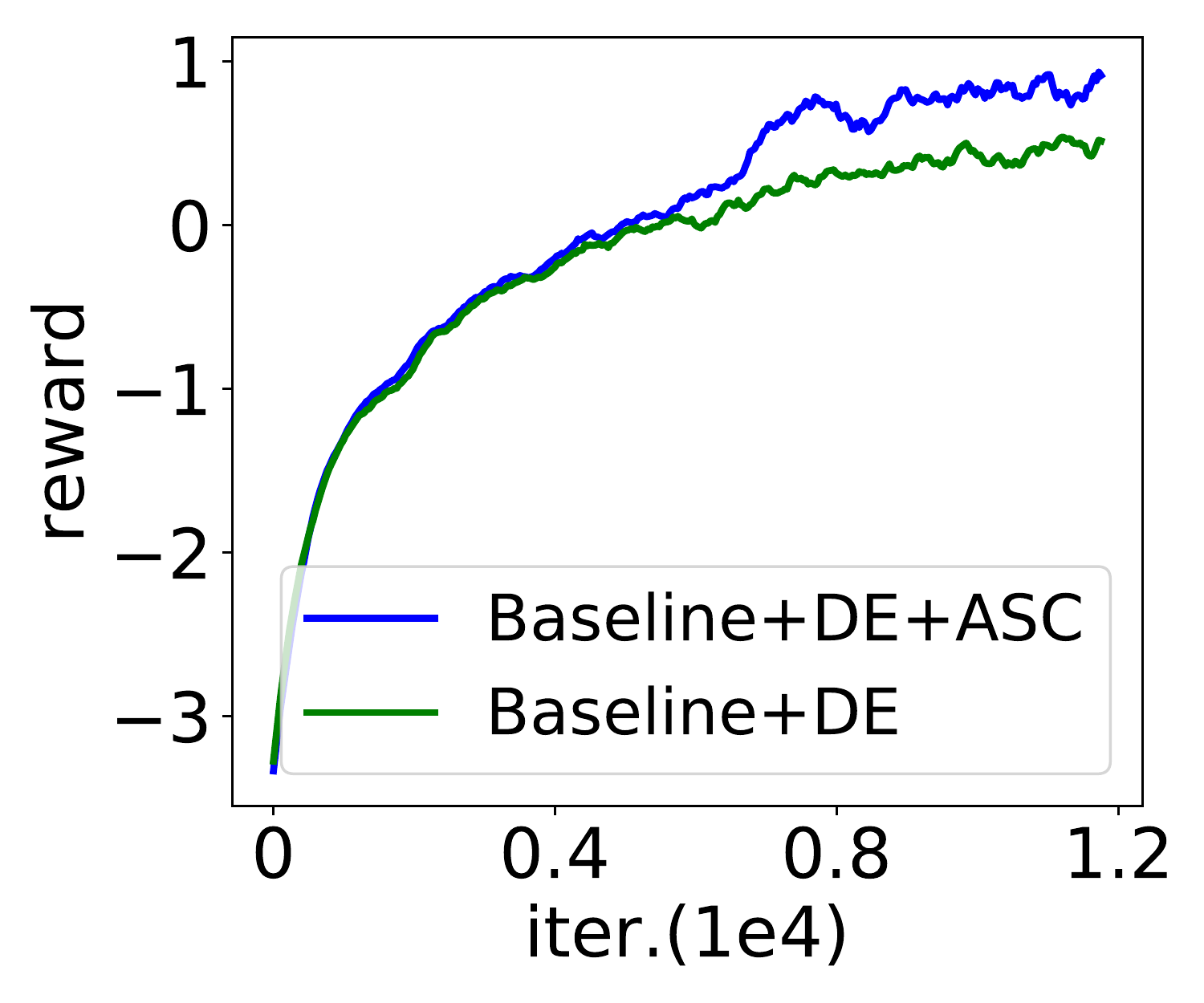}
			\caption{Training rewards}
			\label{fig:rewards2}
		\end{subfigure}
		\caption{
			Training and testing results in our environment. (a) compares the training rewards between our reward function and original reward function applied on same model. (b) shows the policy entropy during training between baseline and model with diversity exploration. (c) compares the training rewards between a model trained by diversity and self-correction and a model with diversity only.
		}
		\label{fig:plots}
		\vspace{-0.7cm}
	\end{figure}

	\noindent\textbf{Baseline} Figure~\ref{fig:plots}(a) illustrates our baseline algorithm trained with different reward functions. The training reward of our baseline increases fast at early steps and converges after 8000 steps. Results suggest that our baseline is sufficiently trained, and our reward function makes the model converge faster and perform better. 
	
\begin{wrapfigure}{r}{0.5\textwidth}
		\vspace{-0.7cm}
		\centering
		\begin{subfigure}[b]{0.24\textwidth}
			\includegraphics[width=\textwidth]{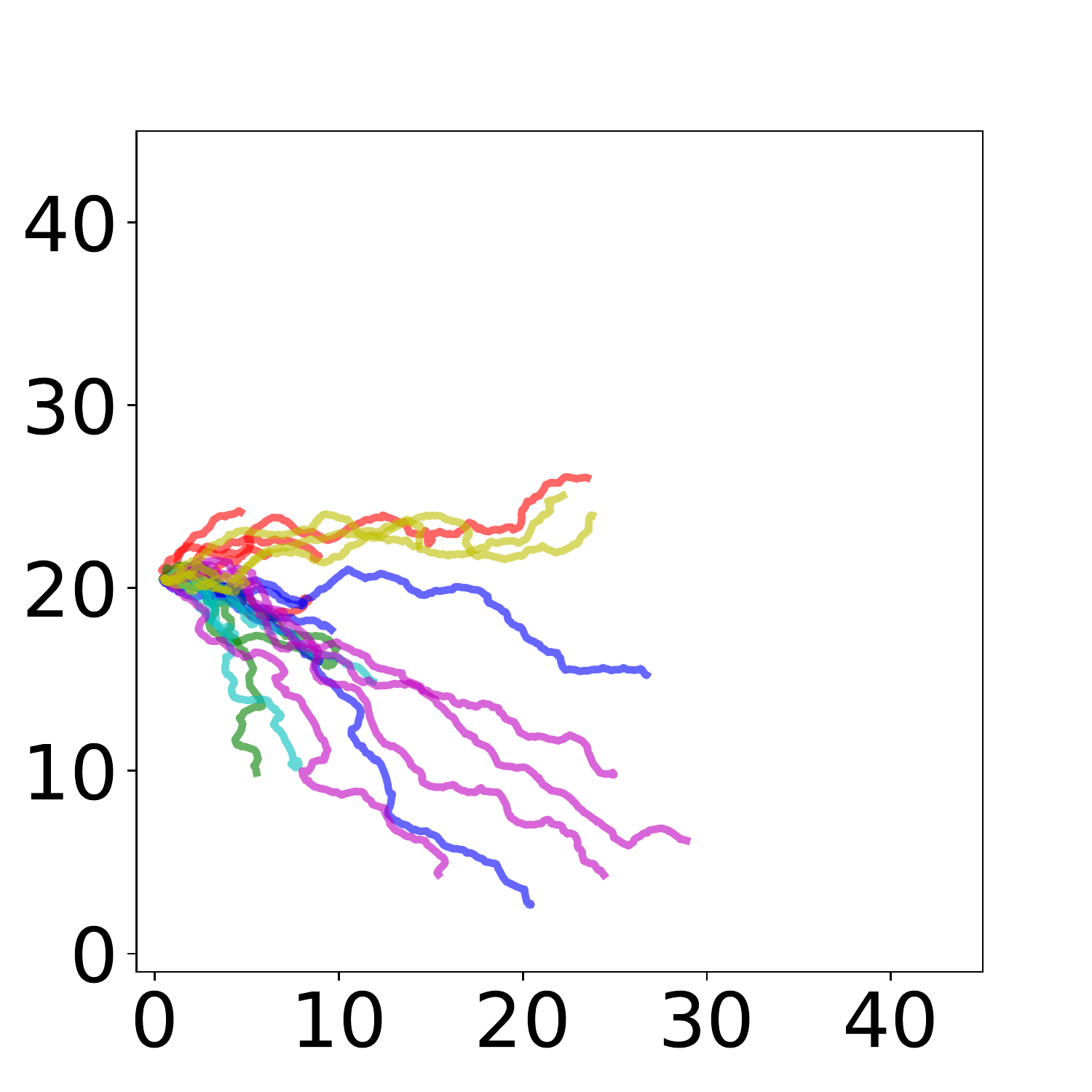}
			\caption{Baseline}
		\end{subfigure}
		\begin{subfigure}[b]{0.24\textwidth}
			\includegraphics[width=\textwidth]{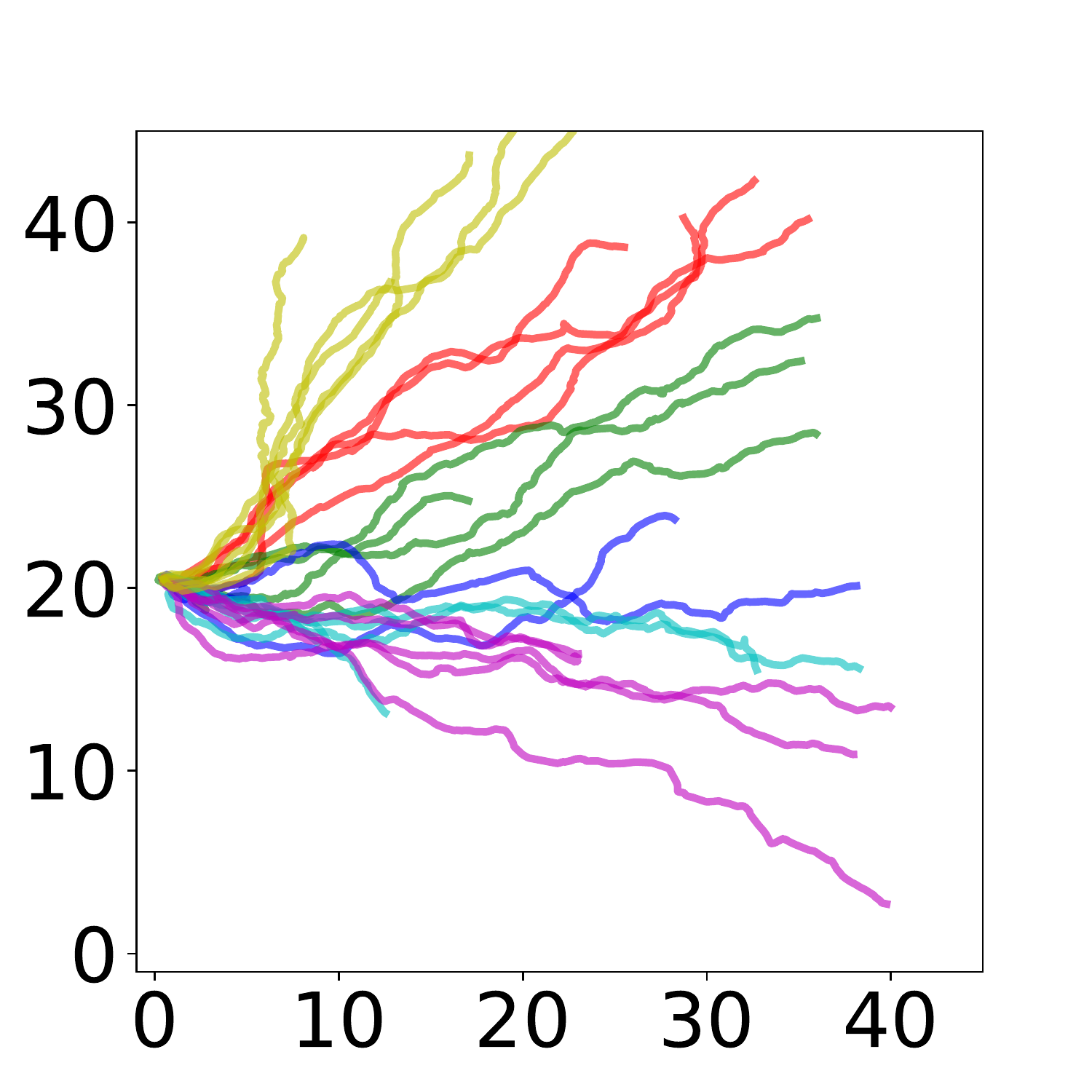}
			\caption{Baseline+DE}
		\end{subfigure}
		\caption{Trajectories sampled from baseline and baseline+DE in different tasks. Same color stands for trajectories sampled from same tasks. }
		\label{fig:diversity}
		\vspace{-0.3cm}
	\end{wrapfigure}

	\noindent\textbf{Diversity Exploration} To understand how diversity exploration benefits our model, we compare the trajectories obtained by sampling from robots that are put in an empty square. We visualize trajectories sampled from same task condition with same color and trajectories sampled from different conditions with different colors. We can see from Figure~\ref{fig:diversity} that trajectories sampled from a model trained by diversity exploration are scattered broadly while trajectories from the same tasks are gathered. It indicates that skills belong to the same class are similar and skills from different classes are distinguishable. 
	By diversity exploration, the model learns to behave differently with respect to different task conditions. The correlation between task information and skills has been successfully increased. 
	
	% \begin{wrapfigure}{r}{0.5\textwidth}
	% 	\begin{center}
	% 		\includegraphics[width=0.45\textwidth]{diversity.pdf}
	% 	\end{center}
	% 	\caption{Plot of trajectory sampled from baseline model and model trained by diversity exploration}
	% 	\label{fig:diversity}
	% \end{wrapfigure}
	
	We also \hui{show} how diversity exploration effects our training process. As Figure~\ref{fig:plots}(b) shows, the two models have no policy entropy difference at the very beginning since policies in the early stages are close to random policy. After learning some skills, the model with diversity exploration has higher policy entropy during the training process. Higher policy entropy encourages the model to explore temporal-action space more sufficiently and try more skills.

	\noindent\textbf{Self-Correction} We evaluate the effectiveness of the self-correction objective. We first investigate how rewards increase in the training process. We compare two models with the same hyperparameter, one uses the self-correction objective to work with diversity exploration while the other is optimized by diversity objective only. Figure~\ref{fig:plots}(c) clearly demonstrates that performance can be significantly improved by self-correction in the latter stages of the training process. This indicates that experience replay is critical since the model can learn from good previous experiences to correct its skills. 
	
	\begin{figure}
		\centering
		\includegraphics[width=0.85\linewidth]{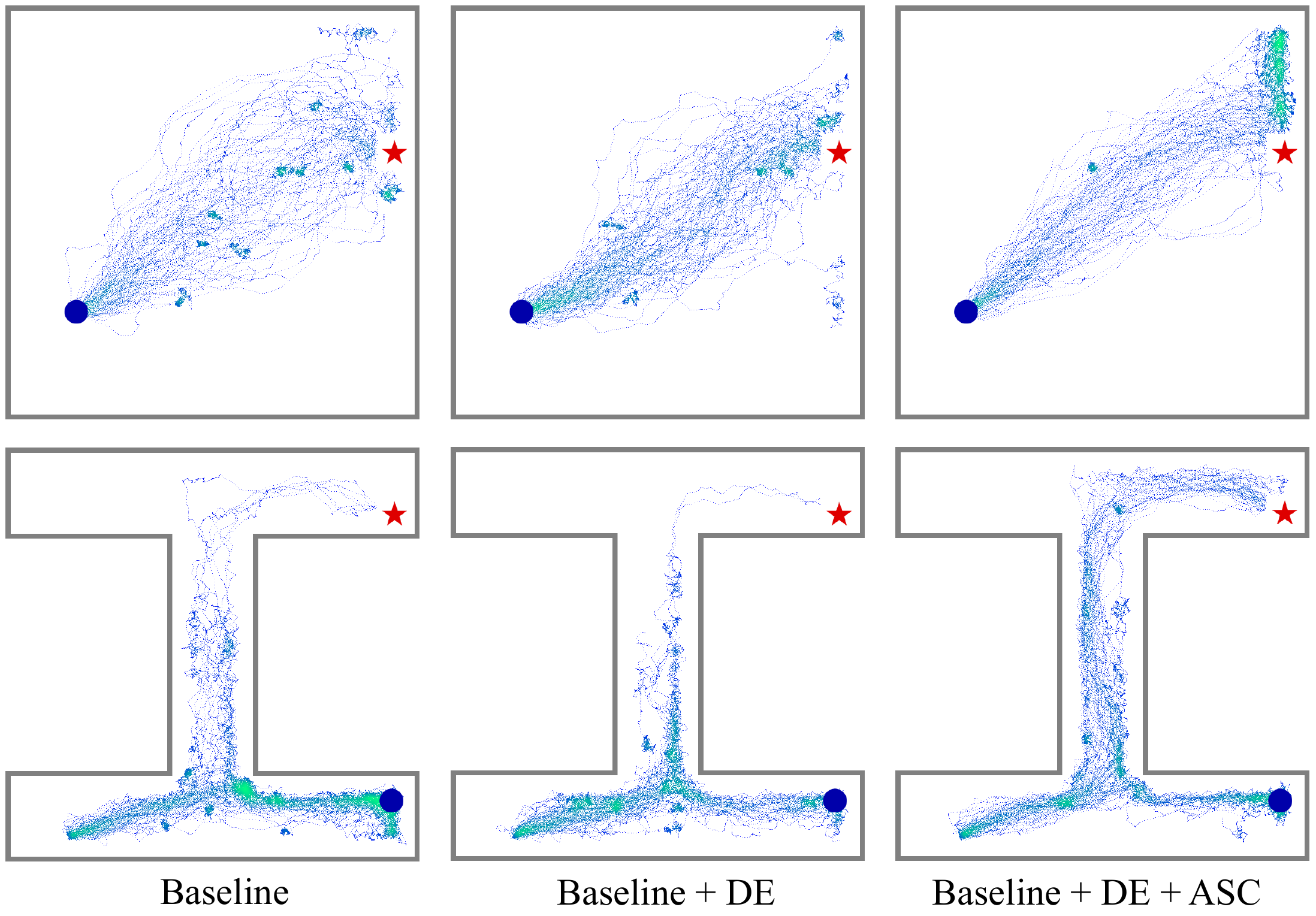}
		\caption{
			The trajectories of different models. We visualize the trajectories of three models in the testing phase. For each maze, the blue point is the starting point, and the red star is the goal point. A robot stops when it "touches" the goal. 
			%, baseline, model with diversity exploration only and model with diversity and self-correction sampled from testing in different mazes. For each maze, blue point is where robot start and red star is the position where goal is. A robot stops when it "touches" the goal square. 
		}
		\label{fig:performance}
		\vspace{-0.5cm}
	\end{figure}
	
	To show how our method benefits from self-correction, we run model in different mazes and visualize the collected trajectories in Figure~\ref{fig:performance}. In the square maze, trajectories from the baseline model are more disorganized. The baseline model does not have a clear mode of behavior so that its trajectories scattered in a large area. By diversity exploration, the model learns to perform a small set of determined skills under the specific task condition, which is why trajectories are more concentrated. 
	For model combined with diversity and self-correction, skills are more certain, and the robot has a higher rate of success. 
	In the second maze, the baseline model and model with diversity tend to get stuck in the lower left corner. Model equipped with self-correction, however, is more likely to go to the right place. 
	
	\begin{table}
		\centering
		\caption {Shorten distance (SD) for each maze and NSD in total. Tested models include baseline, model with diversity exploration (DE) only, model with self-correction (ASC) only and model combined with diversity exploration and self-correction (DE+ASC).}
		\scalebox{0.95} {
			\begin{tabular}{lccccccccc}
				\hline
				\textbf{}   & line & corner\ 1  & corner\ 2 & square\ 1 & square\ 2 & maze\ 1 & maze\ 2 & \% NSD & reward\\
				\hline
				distance & 8.00 & 8.00 & 16.00 & 8.00 & 12.00 & 16.00 & 16.00 & - & -\\
				\hline
				baseline & 5.38 &  5.13 & 10.42 & 5.83 & 9.68 & 2.55 & 4.27 & 57.87 & 1.16 \\
				DE         & 6.16  &  6.24 & 9.28  & 5.99 & 9.55 & 3.84 & 3.26 & 61.50 & 1.44 \\
				SC        & 6.03&\bf{7.03}&11.85&6.99 & 9.85 & 2.61 & 4.66 & 66.41 & 1.57 \\
				DE+SC  & \bf{6.64}   &  6.98 & \bf{12.04} & \bf{7.21} & \bf{10.36} & \bf{4.00} & \bf{10.74} & \bf{76.22} & \bf{1.77}\\
				\hline
		\end{tabular} }
		\label{table:result}
		\vspace{-0.4cm}
	\end{table}
	
	Lastly, we compare shorten distance (SD) for each maze and normalized shorten distance  (NSD) in total with respect to four models, baseline, model with diversity exploration  (DE) only, the model with self-correction (ASC) only and model combined with diversity exploration and self-correction  (DE+ASC). We collect 50 trajectories for every model per maze. From Table~\ref{table:result}, both diversity exploration and self-correction methods benefit the procedure of continual reinforcement learning. Furthermore, the two methods are remarkably reciprocal, especially in complex mazes like maze 1 and maze 2. The model combined with two methods achieves a great improvement and outperforms the rest of them.

	\section{Conclusion}
	
	In this paper, we have proposed the model CDAN to relieve the problem of continuous reinforcement learning in the continuous control domain. Firstly, we consider boosting the correlation between task information and policy skill. Then we introduce the adversarial self-correction method to solve \emph{catastrophic forgetting}. Specifically, we integrate diversity exploration and adversarial self-correction into an end-to-end framework. We demonstrate the effectiveness of each part and how they exploit mutual benefits in our experiments. 
	
	To the best of our knowledge, it is the first work that focuses on continual reinforcement learning in continuous control domain. Due to the limit of time and device, we only conduct experiments on a few tasks with simple maze structures. In the future, we plan to apply our model to more complex environments and more tasks. 
	
	% Even though task information Model needs exploration to adapt

\end{document}